\pdfoutput=1

\documentclass[11pt]{article}

\usepackage{naacl2021}

\usepackage{times}
\usepackage{latexsym}
\usepackage{comment}

\usepackage[T1]{fontenc}

\usepackage[utf8]{inputenc}

\usepackage{microtype}

\usepackage{mathtools,amssymb}
\usepackage{multirow}

\title{Cross-Task Instance Representation Interactions and Label Dependencies \\ for Joint Information Extraction with Graph Convolutional Networks}

\author{Minh Van Nguyen, Viet Dac Lai and Thien Huu Nguyen\\
  Department of Computer and Information Science \\
  University of Oregon, Eugene, Oregon, USA \\
  \texttt{\{minhnv,vietl,thien\}@cs.uoregon.edu}}

\begin{document}
\maketitle
\begin{abstract}
Existing works on information extraction (IE) have mainly solved the four main tasks separately (entity mention recognition, relation extraction, event trigger detection, and argument extraction), thus failing to benefit from inter-dependencies between tasks. This paper presents a novel deep learning model to simultaneously solve the four tasks of IE in a single model (called FourIE). Compared to few prior work on jointly performing four IE tasks, FourIE features two novel contributions to capture inter-dependencies between tasks. First, at the representation level, we introduce an interaction graph between instances of the four tasks that is used to enrich the prediction representation for one instance with those from related instances of other tasks. Second, at the label level, we propose a dependency graph for the information types in the four IE tasks that captures the connections between the types expressed in an input sentence. A new regularization mechanism is introduced to enforce the consistency between the golden and predicted type dependency graphs to improve representation learning. We show that the proposed model achieves the state-of-the-art performance for joint IE on both monolingual and multilingual learning settings with three different languages.
\end{abstract}


\section{Introduction}

Information Extraction (IE) is an important and challenging task in Natural Language Processing (NLP) that aims to extract structured information from unstructured texts. Following the terminology for IE in the popular ACE 2005 program \cite{Walker:05}, we focus on four major IE tasks in this work: entity mention extraction (EME), relation extraction (RE), event trigger detection (ETD), and event argument extraction (EAE).


Given an input sentence, a vast majority of prior work has solved the four tasks in IE independently at both instance and task levels (called independent prediction models). First, at the instance level, each IE task often requires predictions/classifications for multiple instances in a single input sentence. For instance, in RE, one often needs to predict relations for every pair of entity mentions (called relation instances) in the sentence while multiple word spans in the sentence can be viewed as multiple instances where event type predictions have to be made in ETD (trigger instances). As such, most prior work on IE has performed predictions for instances in a sentence separately by treating each instance as one example in the dataset \cite{zhou:05,Nguyen:15b,Santos:15,Chen:15,Nguyen:15a,Lai:20event}. Second, at the task level, prior work on IE tends to perform the four tasks in a pipelined approach where outputs from one task are used as inputs for other tasks (e.g., EAE is followed by EME and ETD) \cite{Li:13,Chen:15,Veyseh:20graph}.

\begin{figure}[t]
\centering
\includegraphics[width=1.04\columnwidth]{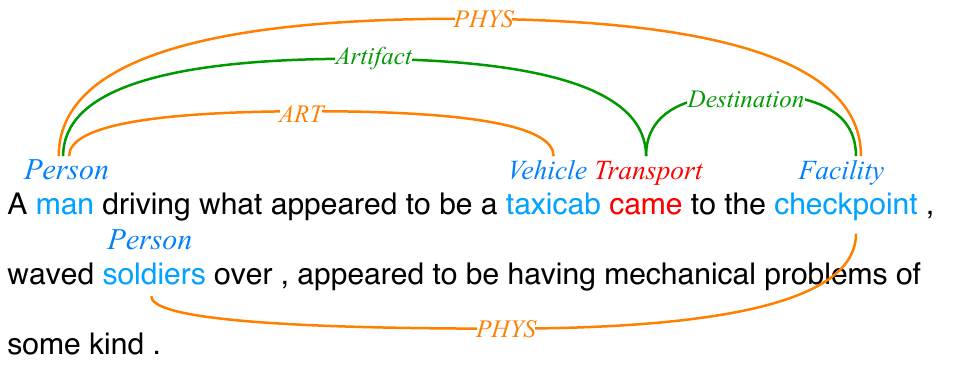}
\caption{A sentence example with the annotations for the four IE tasks. Blue words corresponds to entity mentions while red words are event triggers. Also, orange edges represent relations while green edges indicate argument roles.}
\label{fig:ex}
\end{figure}






\begin{figure*}
    \centering
    \addtolength{\abovecaptionskip}{-2.0mm}
\addtolength{\belowcaptionskip}{-2mm}
    \includegraphics[scale=0.45]{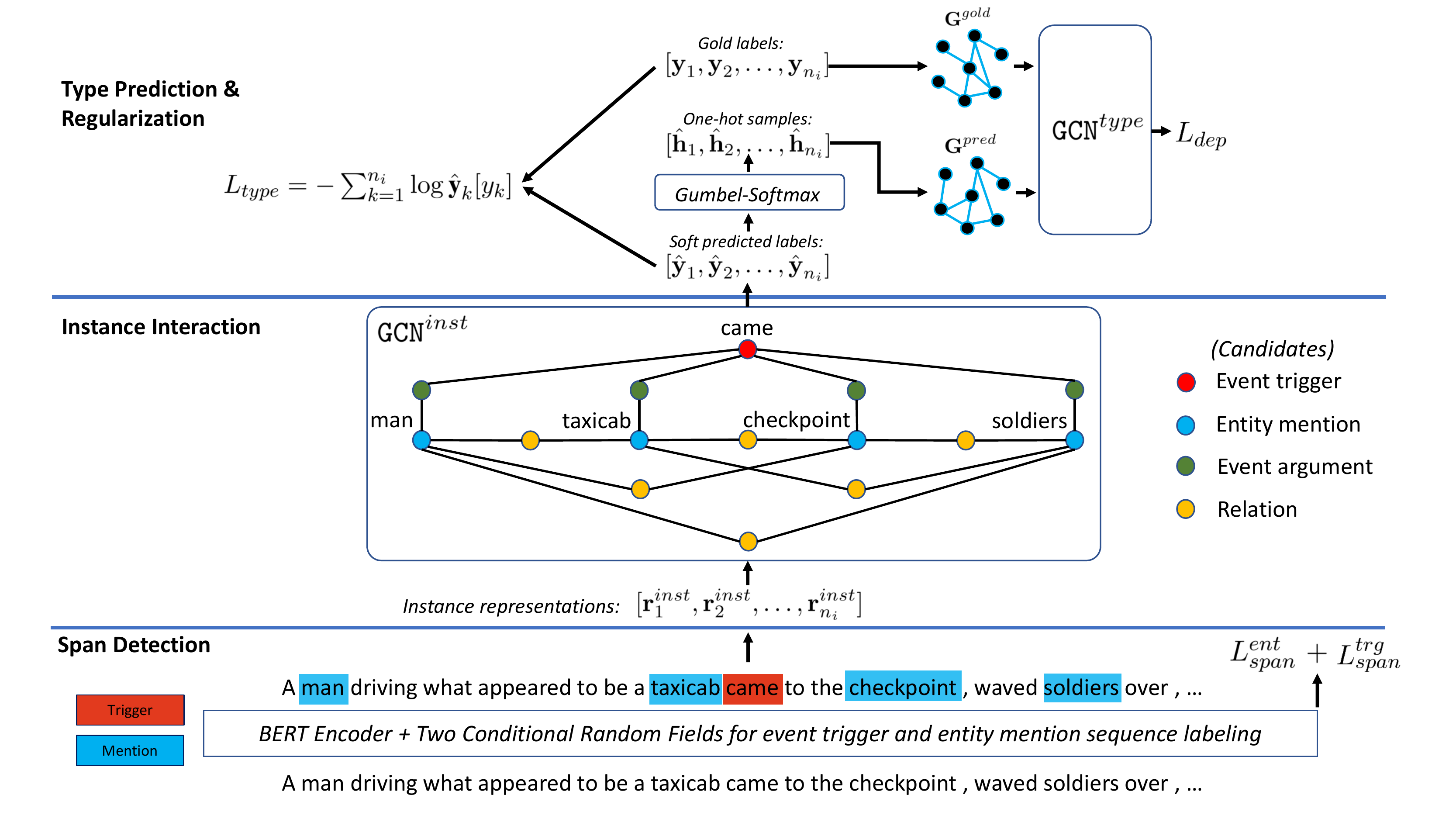}
    \caption{Overall architecture of our proposed model. At the representation level, $\texttt{GCN}^{inst}$ is used to enrich the representations for instances of the four tasks. At the label level, $\texttt{GCN}^{type}$ is responsible for capturing the connections between the types in the dependency graphs, thus helping the model learn the structural difference between the gold graph $\textbf{G}^{gold}$ and the predicted graph $\textbf{G}^{pred}$.}
    \label{fig:system}
\end{figure*}

Despite its popularity, the main issue of the independent prediction models is that they suffer from the error propagation between tasks and the failure to exploit the cross-task and cross-instance inter-dependencies within an input sentence to improve the performance for IE tasks. For instance, such systems are unable to benefit from the dependency that the {\it Victim} of a {\it Die} event has a high chance to also be the {\it Victim} of an {\it Attack} event in the same sentence (i.e., type or label dependencies). To address these issues, some prior work has explored joint inference models where multiple tasks of IE are performed simultaneously for all task instances in a sentence, using both feature-based models \cite{Roth:04,Li:13,Miwa:14,Yang:16} and recent deep learning models \cite{Miwa:16,Zhang:19}. However, such prior work has mostly considered joint models for a subset of the four IE tasks (e.g., EME+RE or ETD+EAE), thus still suffering from the error propagation issue (with the missing tasks) and failing to fully exploit potential inter-dependencies between the four tasks. To this end, this work aims to design a single model to simultaneously solve the four IE tasks for each input sentence (joint four-task IE) to address the aforementioned issues of prior joint IE work.




Few recent work has considered joint four-task IE, using deep learning to produce state-of-the-art (SOTA) performance for the tasks \cite{wadden2019entity,lin2020joint}. However, there are still two problems that hinder further improvement of such models. First, at the instance level, an important component of deep learning models for joint IE involves the representation vectors of the instances that are used to perform the corresponding prediction tasks for IE in an input sentence (called predictive instance representations). For joint four-task IE, we argue that there are inter-dependencies between predictive representation vectors of related instances for the four tasks that should be modeled to improve the performance for IE. For instance, the entity type information encoded in the predictive representation vector for an entity mention can constrain the argument role that the representation vector for a related EAE instance (e.g., involving the same entity mention and some event trigger in the same sentence) should capture and vice versa. As such, prior work for joint four-task IE has only computed predictive representation vectors for instances of the tasks independently using shared hidden vectors from some deep learning layer \cite{wadden2019entity,lin2020joint}. Although this shared mechanism helps capture the interaction of predictive representation vectors to some extent, it fails to explicitly present the connections between related instances of different tasks and encode them into the representation learning process. Consequently, to overcome this issue, we propose a novel deep learning model for joint four-task IE (called {\it FourIE}) that creates a graph structure to explicitly capture the interactions between related instances of the four IE tasks in a sentence. This graph will then be consumed by a graph convolutional network (GCN) \cite{kipf:17,Nguyen:18a} to enrich the representation vector for an instance with those from the related (neighboring) instances for IE.


Second, at the task level, existing joint four-task models for IE have only exploited the cross-task type dependencies in the decoding step to constrain predictions for the input sentence (by manually converting the type dependency graphs of the input sentence into global feature vectors for scoring the predictions in the beam search-based decoding) \cite{lin2020joint}. The knowledge from cross-task type dependencies thus cannot contribute to the training process of the IE models. This is unfortunate as we expect that deeper integration of this knowledge into the training process could provide useful information to enhance representation learning for IE tasks. To this end, we propose to use the knowledge from cross-task type dependencies to obtain an additional training signal for each sentence to directly supervise our joint four-task IE model. In particular, our motivation is that the types expressed in a sentence for the four IE tasks can be organized into a dependency graph between the types (global type dependencies for the sentence). As such, in order for a joint model to perform well, the type dependency graph generated by its predictions for a sentence should be similar to the dependency graph obtained from the golden types (i.e., a global type constraint on the predictions in the training step). A novel regularization term is thus introduced into the training loss of our joint model to encode this constraint, employing another GCN to learn representation vectors for the predicted and golden dependency graphs to facilitate the graph similarity promotion. To our knowledge, this is the first work that employs global type dependencies to regularize joint models for IE.

Finally, our extensive experiments demonstrate the effectiveness of the proposed model on benchmark datasets in three different languages (e.g., English, Chinese, and Spanish), leading to state-of-the-art performance on different settings.

\section{Problem Statement and Background}


{\bf Problem Statement}: The joint four-task IE problem in this work takes a sentence as the input and aims to jointly solve four tasks EAE, ETD, RE, and EAE using an unified model. As such, the goal of EME is to detect and classify entity mentions (names, nominals, pronouns) according to a set of predefined (semantic) entity types (e.g., {\it Person}). Similarly, ETD seeks to identify and classify event triggers (verbs or normalization) that clearly evoke an event in some predefined set of event types (e.g., {\it Attack}). Note that event triggers can involve multiple words. For RE, its concern is to predict the semantic relationship between two entity mentions in the sentence. Here, the set of relations of interest is also predefined and includes a special type of {\it None} to indicate {\it no-relation}. Finally, in EAE, given an event trigger, the systems need to predict the roles (also in a predefined set with a special type {\it None}) that each entity mention plays in the corresponding event. Entity mentions are thus also called event argument candidates in this work. Figure \ref{fig:ex} presents a sentence example where the expected outputs for each IE task are illustrated.







\noindent {\bf Graph Convolutional Networks (GCN)}: As GCNs are used extensively in our model, we present their computation process in this section to facilitate the discussion. Given a graph $\textbf{G}=(\textbf{V},\textbf{E})$ where $\textbf{V}=\{v_1,\ldots,v_u\}$ is the node set (with $u$ nodes) and $\textbf{E}$ is the edge set. In GCN, the edges in $\textbf{G}$ are often captured via the adjacency matrix $\textbf{A} \in \mathbb{R}^{u\times u}$. Also, each node $v_i \in \textbf{V}$ is associated with an initial hidden vector $\textbf{v}^0_i$. As such, a GCN model involves multiple layers of abstraction in which the hidden vector $\textbf{v}^l_i$ for the node $v_i \in \textbf{V}$ at the $l$-th layer is computed by ($l\ge 1$):
\begin{equation*}
\small
    \textbf{v}^l_i = \textrm{ReLU}(\frac{\sum^u_{j=1} \textbf{A}_{ij}\textbf{W}^l\textbf{v}^{l-1}_j + \textbf{b}^l}{\sum^u_{j=1} \textbf{A}_{ij}} )
\end{equation*}
where $\textbf{W}^l$ and $\textbf{b}^l$ are trainable weight and bias at the $l$-th layer. Assuming $N$ GCN layers, the hidden vectors for the nodes in $\textbf{V}$ at the last layer $\textbf{v}^N_1,\ldots,\textbf{v}^N_u$ would capture richer and more abstract information for the nodes, serving as the outputs of the GCN model. This process is denoted by: $\textbf{v}^N_1,\ldots,\textbf{v}^N_u = \texttt{GCN}(\textbf{A}; \textbf{v}^0_1,\ldots,\textbf{v}^0_u; N)$.


\section{Model}

Given an input sentence $\textbf{w}=[w_1, w_2,\ldots, w_n]$ (with $n$ words), our model for joint four-task IE on $\textbf{w}$ involves three major components: (i) Span Detection, (ii) Instance Interaction, and (iii) Type Dependency-based Regularization.


\subsection{Span Detection}

This component aims to identify spans of entity mentions and event triggers in $\textbf{w}$ that would be used to form the nodes in the interaction graph between different instances of our four IE tasks for $\textbf{w}$. As such, we formulate the span detection problems as sequence labeling tasks where each word $w_i$ in $\textbf{w}$ is associated with two BIO tags to capture the span information for entity mentions and event triggers in $\textbf{w}$. Note that we do not predict entity types and event types at this step, leading to only three possible values (i.e., B, I, and O) for the tags of the words.



In particular, following \cite{lin2020joint}, we first feed $\textbf{w}$ into the pre-trained BERT encoder \cite{devlin2018bert} to obtain a sequence of vectors $\textbf{X}=[\textbf{x}_1, \textbf{x}_2,\ldots, \textbf{x}_n]$ to represent $\textbf{w}$. Here, each vector $\textbf{x}_i$ serves as the representation vector for the word $w_i \in \textbf{w}$ that is obtained by averaging the hidden vectors of the word-pieces of $w_i$ returned by BERT. Afterward, $\textbf{X}$ is fed into two conditional random field (CRF) layers to determine the best BIO tag sequences for event mentions and event triggers for $\textbf{w}$, following \cite{chiu2016named}. As such, the Viterbi algorithm is used to decode the input sentence while the negative log-likelihood losses are employed as the training objectives for the span detection component of the model. For convenience, let $L^{ent}_{span}$ and $L^{trg}_{span}$ be the negative log-likelihoods of the gold tag sequences for entity mentions and event triggers (respectively) for $\textbf{w}$. These terms will be included in the overall loss function of the model later.

\subsection{Instance Interaction}


Based on the tag sequences for $\textbf{w}$ from the previous component, we can obtain two separate span sets for the entity mentions and event triggers in $\textbf{w}$ (the golden spans are used in the training phase to avoid noise). For the next computation, we first compute a representation vector for each span $(i,j)$ ($1 \le i \le j \le n$) in these two sets by averaging the BERT-based representation vectors for the words in this span (i.e., $\textbf{x}_i,\ldots,\textbf{x}_j$). For convenience, let $\textbf{R}^{ent}=\{\textbf{e}_1, \textbf{e}_2,\ldots, \textbf{e}_{n_{ent}}\}$ ($n_{ent} = |\textbf{R}^{ent}|$) and $\textbf{R}^{trg}=\{\textbf{t}_1, \textbf{t}_2,\ldots,\textbf{t}_{n_{trg}}\}$ ($n_{trg} = |\textbf{R}^{trg}|$) be the sets of span representation vectors for the entity mentions and event triggers in $\textbf{w}$\footnote{We will also refer to entity mentions and event triggers interchangeably with their span representations $\textbf{e}_i$ and $\textbf{t}_i$ in this work.}. The goal of this component is to leverage such span representation vectors to form instance representations and enrich them with instance interactions to perform necessary predictions in IE.

\noindent \textbf{Instance Representation}. Prediction instances in our model amount to the specific objects that we need to predict a type for one of the four IE tasks. As such, the prediction instances for EME and ETD, called entity and trigger instances, correspond directly to the entity mentions and event triggers in $\textbf{R}^{ent}$ and $\textbf{R}^{trg}$ respectively (as we need to predict the entity types for $\textbf{e}_i \in \textbf{R}^{ent}$ and the event types for $\textbf{t}_i \in \textbf{R}^{trg}$ in this step). Thus, we also use $\textbf{R}^{ent}$ and $\textbf{R}^{trg}$ as the sets of initial representation vectors for the entity/event instances for EME and ETD in the following. Next, for RE, the prediction instances (called relation instances) involve pairs of entity mentions in $\textbf{R}^{ent}$. To obtain the initial representation vector for a relation instance, we concatenate the representation vectors of the two corresponding entity mentions, leading to the set of representation vectors $\textbf{rel}_{ij}$ for relation instances: $\textbf{R}^{rel} = \{\textbf{rel}_{ij}=[\textbf{e}_i, \textbf{e}_j] \mid \textbf{e}_i, \textbf{e}_j \in \textbf{R}^{ent}, i < j\}$ ($|\textbf{R}^{rel}| = n_{ent}(n_{ent}-1)/2$). Finally, for EAE, we form the prediction instances (called argument instances) by pairing each event trigger in $\textbf{R}^{trg}$ with each entity mention in $\textbf{R}^{ent}$ (for the argument role predictions of the entity mentions with respect to the event triggers/mentions). By concatenating the representation vectors of the paired entity mentions and event triggers, we generate the initial representation vectors $\textbf{arg}_{ij}$ for the corresponding argument instances: $\textbf{R}^{arg} = \{\textbf{arg}_{ij}=[\textbf{t}_i, \textbf{e}_j] \mid \textbf{t}_i \in \textbf{R}^{trg}, \textbf{e}_j \in \textbf{R}^{ent}\}$ ($|\textbf{R}^{arg}| = n_{trg} n_{ent}$)\footnote{In our implementation, $\textbf{R}^{rel}$ and $\textbf{R}^{arg}$ are transformed into vectors of the same size with those in $\textbf{R}^{ent}$ and $\textbf{R}^{trg}$ (using one-layer feed forward networks) for future computation.}. We also use the prediction instances and their representation vectors interchangeably in this work.

\noindent \textbf{Instance Interaction}. The initial representation vectors for the instances so far do not explicitly consider beneficial interactions between related instances. To address this issue, we explicitly create an interaction graph between the prediction instances for the four IE tasks to connect related instances to each other. This graph will be consumed by a GCN model to enrich instance representations with interaction information afterward. In particular, the node set $\textbf{N}^{inst}$ in our instance interaction graph $\textbf{G}^{inst} = \{\textbf{N}^{inst},\textbf{E}^{inst}\}$ involves all prediction instances for the four IE tasks, i.e., $\textbf{N}^{inst} = \textbf{R}^{ent} \cup \textbf{R}^{trg} \cup \textbf{R}^{rel} \cup \textbf{R}^{arg}$. The edge set $\textbf{E}^{inst}$ then captures instance interactions by connecting the instance nodes in $\textbf{N}^{inst}$ that involve the same entity mentions or event triggers (i.e., two instances are related if they concern the same entity mention or event trigger). As such, the edges in $\textbf{E}^{inst}$ are created as follows:

(i) An entity instance node $\textbf{e}_i$ is connected to all relation instance nodes of the forms $\textbf{rel}_{ij} = [\textbf{e}_i, \textbf{e}_j]$ and $\textbf{rel}_{ki} = [\textbf{e}_k, \textbf{e}_i]$ (sharing entity mention $\textbf{e}_i$).


(ii) An entity instance node $\textbf{e}_j$ is connected to all argument instance nodes of the form $\textbf{arg}_{ij} = [\textbf{t}_i, \textbf{e}_j]$ (sharing entity mention $\textbf{e}_j$).


(iii) A trigger node $\textbf{t}_i$ is connected to all argument instance nodes of the form $\textbf{arg}_{ij} = [\textbf{t}_i, \textbf{e}_j]$ (i.e., sharing event trigger $\textbf{t}_i$).




\noindent \textbf{GCN}. To enrich the representation vector for an instance in $\textbf{N}^{inst}$ with the information from the related (neighboring) nodes, we feed $\textbf{G}^{inst}$ into a GCN model (called $\texttt{GCN}^{inst}$). For convenience, we rename the initial representation vectors of all the instance nodes in $\textbf{N}^{inst}$ by: $\textbf{N}^{inst} = \{\textbf{r}_1,\ldots,\textbf{r}_{n_i}\}$ ($n_i = |\textbf{N}^{inst}|$). Also, let $\textbf{A}^{inst} \in \{0,1\}^{n_i \times n_i}$ be the adjacency matrix of the interaction graph $\textbf{G}^{inst}$ where $\textbf{A}^{inst}_{ij} = 1$ if the instance nodes $\textbf{r}_i$ and $\textbf{r}_j$ are connected in $\textbf{G}^{inst}$ or $i = j$ (for self-connections). The interaction-enriched representation vectors for the instances in $\textbf{N}^{inst}$ are then computed by the $\texttt{GCN}^{inst}$ model: $\textbf{r}^{inst}_1,\ldots,\textbf{r}^{inst}_{n_i} = \texttt{GCN}^{inst}(\textbf{A}^{inst}; \textbf{r}_1,\ldots,\textbf{r}_{n_i}; N_i)$ where $N_i$ is the number of layers for the $\texttt{GCN}^{inst}$ model.

\noindent \textbf{Type Embedding and Prediction}. Finally, the enriched instance representation vectors $\textbf{r}^{inst}_1,\ldots,\textbf{r}^{inst}_{n_i}$ will be used to perform the predictions for the four IE tasks. In particular, let $t_k \in \{ent, trg, rel, arg\}$ be the corresponding task index and $y_k$ be the ground-truth type (of the task $t_k$) for the prediction instance $\textbf{r}_k$ in $\textbf{N}^{inst}$. Also, let $\mathcal{T} = \mathcal{T}^{ent} \cup \mathcal{T}^{trg} \cup \mathcal{T}^{rel} \cup \mathcal{T}^{arg}$ be the union of the possible entity types (in $\mathcal{T}^{ent}$ for EME), event types (in $\mathcal{T}^{trg}$ for ETD), relations (in $\mathcal{T}^{rel}$ for RE), and argument roles (in $\mathcal{T}^{arg}$ for EAE) in our problem ($y_k \in \mathcal{T}^{t_k}$). Note that $\mathcal{T}^{rel}$ and $\mathcal{T}^{arg}$ contain the special types {\it None}. To prepare for the type predictions and the type dependency modeling in the next steps, we associate each type in $\mathcal{T}$ with an embedding vector (of the same size as $\textbf{e}_i$ and $\textbf{t}_i$) that is initialized randomly and updated during our training process. For convenience, let $\mathcal{T} = [\bar{\textbf{t}}_1,\ldots,\bar{\textbf{t}}_{n_t}]$ where $\bar{\textbf{t}}_i$ is used interchangeably for both a type and its embedding vector in $\mathcal{T}$ ($n_t$ is the total number of types). As such, to perform the prediction for an instance $\textbf{r}_k$ in $\textbf{N}^{inst}$, we compute the dot products between $\textbf{r}^{inst}_k$ and each type embedding vectors in $\mathcal{T}^{t_k}\cap \mathcal{T}$ to estimate the possibilities that $\textbf{r}_k$ has a type in $\mathcal{T}^{t_k}$. Afterward, these scores are normalized by the softmax function to obtain the probability distribution $\hat{\textbf{y}}_k$ over the possible types in $\mathcal{T}^{t_k}$ for $\textbf{r}_k$: $\hat{\textbf{y}}_k = softmax(\textbf{r}^{inst}_k \bar{\textbf{t}}^T | \bar{\textbf{t}} \in \mathcal{T}^{t_k} \cap \mathcal{T})$. In the decoding phase, the predicted type $\hat{y}_k$ for $\textbf{r}_k$ is obtained via the \texttt{argmax} function (greedy decoding): $\hat{y}_k = \texttt{argmax } \hat{\textbf{y}}_k$. The negative log-likelihood over all the prediction instances is used to train the model: $L_{type} = -\sum^{n_i}_{k=1} \log \hat{\textbf{y}}_k[y_k]$.

\subsection{Type Dependency-based Regularization}

In this section, we aim to obtain the type dependencies across tasks and use them to supervise the model in the training process (to improve the representation vectors for IE). As presented in the introduction, our motivation is to generate global dependency graphs between types of different IE tasks for each input sentence whose representations are leveraged to regularize the model during training. In particular, starting with the golden types $\textbf{y} = y_1,y_2,\ldots,y_{n_i}$ and the predicted types $\hat{\textbf{y}} = \hat{y}_1,\hat{y}_2,\ldots,\hat{y}_{n_i}$ for the instance nodes in $\textbf{N}^{inst}$, we build two dependency graphs $\textbf{G}^{gold}$ and $\textbf{G}^{pred}$ to capture the global type dependencies for the tasks (called the golden and predicted dependency graphs respectively). Afterward, to supervise the training process, we seek to constrain the model so the predicted dependency graph $\textbf{G}^{pred}$ is similar to the golden graph $\textbf{G}^{gold}$ (i.e., using the dependency graphs as the bridges to inject the global type dependency knowledge in $\textbf{G}^{gold}$ into the model).


\noindent \textbf{Dependency Graph Construction}. Both $\textbf{G}^{gold}$ and $\textbf{G}^{pred}$ involve the types of all the four IE tasks in $\mathcal{T}$ as the nodes. To encode the type dependencies, the connections/edges in $\textbf{G}^{gold}$ are computed based on the golden types $\textbf{y} = y_1,y_2,\ldots,y_{n_i}$ for the instance nodes in $\textbf{N}^{inst}$ as follows:


(i) For each relation instance node $\textbf{r}_k =[\textbf{e}_i, \textbf{e}_j] \in \textbf{N}^{inst}$ that has the golden type $y_k \ne None$, the relation type node $y_k$ is connected to the nodes of the golden entity types for the corresponding entity mentions $\textbf{e}_i$ and $\textbf{e}_j$ (called {\bf entity\_relation type edges}).


(ii) For each argument instance node $\textbf{r}_k =[\textbf{t}_i, \textbf{e}_j]$ that has the role type $y_k \ne None$, the role type node $y_k$ is connected to both the node for the golden event type of $\textbf{t}_i$ (called {\bf event\_argument type edges}) and the node for the golden entity type of $\textbf{e}_j$ (called {\bf entity\_argument type edges}).

The same procedure can be applied to build the predicted dependency graph $\textbf{G}^{pred}$ based on the predicted types $\hat{\textbf{y}} = \hat{y}_1,\hat{y}_2,\ldots,\hat{y}_{n_i}$. Also, for convenience, let $\textbf{A}^{gold}$ and $\textbf{A}^{pred}$ (of size $n_t \times n_t$) be the binary adjacency matrices of $\textbf{G}^{gold}$ and $\textbf{G}^{pred}$ (including the self-loops) respectively.

\noindent \textbf{Regularization}. In the next step, we obtain the representation vectors for the dependency graphs $\textbf{G}^{gold}$ and $\textbf{G}^{pred}$ by feeding them into a GCN model (called $\texttt{GCN}^{type}$). This GCN model has $N_t$ layers and uses the initial type embeddings $\mathcal{T} = [\bar{\textbf{t}}_1,\ldots,\bar{\textbf{t}}_{n_t}]$ as the inputs. In particular, the outputs of $\texttt{GCN}^{type}$ for the two graphs involve $\bar{\textbf{t}}^{gold}_1,\ldots,\bar{\textbf{t}}^{gold}_{n_t} = \texttt{GCN}^{type}(\textbf{A}^{gold}; \bar{\textbf{t}}_1,\ldots,\bar{\textbf{t}}_{n_t}; N_t)$ and $\bar{\textbf{t}}^{pred}_1,\ldots,\bar{\textbf{t}}^{pred}_{n_t} = \texttt{GCN}^{type}(\textbf{A}^{pred}; \bar{\textbf{t}}_1,\ldots,\bar{\textbf{t}}_{n_t}; N_t)$ that encode the underlying information for the type dependencies presented in $\textbf{G}^{gold}$ and $\textbf{G}^{pred}$. Finally, to promote the similarity of the type dependencies in $\textbf{G}^{gold}$ and $\textbf{G}^{pred}$, we introduce the mean square difference between their $\texttt{GCN}^{type}$-induced representation vectors into the overall loss function for minimization: $L_{dep} = \sum^{n_t}_{i=1} ||\bar{\textbf{t}}^{gold}_i - \bar{\textbf{t}}^{pred}_i||^2_2$.

Our final training loss is thus: $L = L^{ent}_{span} + L^{trg}_{span} + L_{type} + \lambda L_{dep}$ ($\lambda$ is a trade-off parameter).


\noindent \textbf{Approximating $\textbf{A}^{pred}$}. We distinguish two types of parameters in our model so far, i.e., the parameters used to compute instance representations, e.g., those in BERT and $\textbf{G}^{inst}$ (called $\theta^{inst}$), and the parameters for type dependency regularization, i.e., those for the type embeddings $\bar{\textbf{t}}_1,\ldots,\bar{\textbf{t}}_{n_t}$ and $\textbf{G}^{type}$ (called $\theta^{dep}$). As such, the current implementation only enables the training signal from $L_{dep}$ to back-propagate to the parameters $\theta^{dep}$ and disallows $L_{dep}$ to influence the instance representation-related parameters $\theta^{inst}$. To enrich the instance representation vectors with type dependency information, we expect $L_{dep}$ to be deeper integrated into the model by also contributing to $\theta^{inst}$. To achieve this goal, we note that the block of back-propagation between $L_{dep}$ and $\theta^{inst}$ is due to their only connection in the model via the adjacency matrix $\textbf{A}^{pred}$, whose values are either one or zero. As such, the values in $\textbf{A}^{pred}$ are not directly dependent on any parameter in $\theta^{inst}$, making it impossible for the back-propagation to flow. To this end, we propose to approximate $\textbf{A}^{pred}$ with a new matrix $\hat{\textbf{A}}^{pred}$ that directly involves $\theta^{inst}$ in its values. In particular, let $\textbf{I}^{inst}$ be the index set of the non-zero cells in $\textbf{A}^{pred}$: $\textbf{I}^{inst} = \{(i,j) | \textbf{A}^{pred}_{ij} = 1\}$. As the elements in $\textbf{I}^{inst}$ are determined by the indexes $i_1,\ldots,i_{n_i}$ in $\mathcal{T}$ of the predicted types $\hat{y}_1,\hat{y}_2,\ldots,\hat{y}_{n_i}$ (respectively), we also seek to compute the values for the approximated matrix $\hat{\textbf{A}}^{pred}$ based on such indexes. Accordingly, we first define the matrix $\textbf{B} = \{b_{ij}\}_{i,j=1..n_t}$ where the element $b_{ij}$ at the $i$-th row and $j$-th column is set to $b_{ij} = i*n_t + j$. The approximated matrix $\hat{\textbf{A}}^{pred}$ is then obtained by:
\begin{equation}
\small
\hat{\textbf{A}}^{pred} = \sum_{(i,j) \in \textbf{I}^{inst}} \exp (-\beta(\textbf{B} - i n_t - j)^2)
\end{equation}
Here, $\beta > 0$ is a large constant. For each element $(i,j) \in \textbf{I}^{inst}$, all the elements in the matrix $(\textbf{B} - i n_t - j)^2$ are strictly positive, except for the element at $(i,j)$, which is zero. Thus, with a large value for $\beta$, the matrix $\exp (-\beta(\textbf{B} - i n_t - j)^2)$ has the value of one at cell $(i,j)$ and nearly zero at other cells. Consequently, the values of $\hat{\textbf{A}}^{pred}$ at the positions in $\textbf{I}^{inst}$ are close to one while those at other positions are close to zero, thus approximating our expected matrix $\textbf{A}^{pred}$ and still directly depending on the indexes $i_1,\ldots,i_{n_t}$.


\noindent \textbf{Addressing the Discreteness of Indexes}. Even with the approximation $\hat{\textbf{A}}^{pred}$, the back-propagation still cannot flow from $L_{dep}$ to $\theta^{inst}$ due to the block of the discrete and non-differentiable index variables $i_1,\ldots,i_{n_t}$. To address this issue, we propose to apply the Gumbel-Softmax distribution \cite{jang2016categorical} that enables the optimization of models with discrete random variables, by providing a method to approximate one-hot vectors sampled from a categorical distribution with continuous ones.



In particular, we first rewrite each index $i_k$ by: $i_k = \textbf{h}_k\textbf{c}^T_k$, where $\textbf{c}_k$ is a vector whose each dimension contains the index of a type in $\mathcal{T}^{t_k}$ in the joint type set $\mathcal{T}$, and $\textbf{h}_k$ is the binary one-hot vector whose dimensions correspond to the types in $\mathcal{T}^{t_k}$. $\textbf{h}_k$ is only turned on at the position corresponding to the predicted type $\hat{y}_k \in \mathcal{T}^{t_k}$ (indexed at $i_k$ in $\mathcal{T}$). In our current implementation, $\hat{y}_k$ (thus the index $i_k$ and the one-hot vector $\textbf{h}_k$) is obtained via the \texttt{argmax} function: $\hat{y}_k = \texttt{argmax } \hat{\textbf{y}}_k $, which causes the discreteness. As such, the Gumbel-Softmax distribution method helps to relax \texttt{argmax} by approximating $\textbf{h}_k$ with a sample $\hat{\textbf{h}}_k = \hat{h}_{k,1},\ldots,\hat{h}_{k,|\mathcal{T}^{t_k}|}$ from the Gumbel-Softmax distribution:
\begin{equation}
\small
    \hat{h}_{k,j} = \frac{\textrm{exp}((\textrm{log}(\pi_{k,j}) + g_j)/\tau)}{\sum^{|\mathcal{T}^{t_k}|}_{j'=1} \textrm{exp}((\textrm{log}(\pi_{k,j'}) + g_{j'})/\tau)}
\end{equation}
where $\pi_{k,j} = \hat{\textbf{y}}_{k,j} = softmax_j(\textbf{r}^{inst}_k \bar{\textbf{t}}^T | \bar{\textbf{t}} \in \mathcal{T}^{t_k} \cap \mathcal{T})$, $g_1,\ldots, g_{|\mathcal{T}^{t_k}|}$ are the i.i.d samples drawn from Gumbel(0,1) distribution \cite{gumbel1948statistical}: $g_j = - \textrm{log}(-\textrm{log}(u_j))$ ($u_j \sim \textrm{Uniform}(0,1)$), and $\tau$ is the temperature parameter. As $\tau \rightarrow 0$, the sample $\hat{\textbf{h}}_k$ would become close to our expected one-hot vector $\textbf{h}_k$. Finally, we replace $\textbf{h}_k$ with the approximation $\hat{\textbf{h}}_k$ in the computation for $i_k$: $i_k = \hat{\textbf{h}}_k\textbf{c}^T_k$ that directly depends on $\textbf{r}^{inst}_k$ and is applied in $\hat{\textbf{A}}^{pred}$. This allows the gradients to flow from $L_{dep}$ to the parameters $\theta^{inst}$ and completes the description of our model.

\section{Experiments}

\noindent \textbf{Datasets}. Following the prior work on joint four-task IE \cite{wadden2019entity,lin2020joint}, we evaluate our joint IE model (FourIE) on the ACE 2005 \cite{Walker:05} and ERE datasets that provide annotation for entity mentions, event triggers, relations, and argument roles. In particular, we use three different versions of the ACE 2005 dataset that focus on three major joint inference settings for IE: (i) {\bf ACE05-R} for joint inference of EME and RE, (ii) {\bf ACE05-E} for joint inference of EME, ETD and EAE, and (iii) {\bf ACE05-E+} for joint inference of the four tasks EME, ETD, RE, and EAE. ACE05-E+ is our main evaluation setting as it fits to our model design with the four IE tasks of interest.

\begin{table}[ht]
\centering
\small
\resizebox{0.48\textwidth}{!}{
\begin{tabular}{|l|l|c|c|c|c|}
\hline
\multicolumn{1}{|c|}{\textbf{Datasets}} & \multicolumn{1}{c|}{\textbf{Split}} & \textbf{sents} & \textbf{ents} & \textbf{rels} & \textbf{events} \\ \hline
\multirow{3}{*}{ACE05-R}                & Train                               & 10,051          & 26,473         & 4,788          & -                \\ \cline{2-6} 
                                        & Dev                                 & 2,424           & 6,362          & 1,131          & -                \\ \cline{2-6} 
                                        & Test                                & 2,050           & 5,476          & 1,151          & -                \\ \hline
\multirow{3}{*}{ACE05-E}                & Train                               & 17,172          & 29,006         &  4,664              & 4,202            \\ \cline{2-6} 
                                        & Dev                                 & 923             & 2,451          & 560              & 450              \\ \cline{2-6} 
                                        & Test                                & 832             & 3,017          & 636              & 403              \\ \hline
\multirow{3}{*}{ACE05-E+}               & Train                               & 19,240          & 47,525         & 7,152          & 4,419            \\ \cline{2-6} 
                                        & Dev                                 & 902             & 3,422          & 728            & 468              \\ \cline{2-6} 
                                        & Test                                & 676             & 3,673          & 802            & 424              \\ \hline
\multirow{3}{*}{ERE-EN}                 & Train                               & 14,219          & 38,864         & 5,045          & 6,419            \\ \cline{2-6} 
                                        & Dev                                 & 1,162           & 3,320          & 424            & 552              \\ \cline{2-6} 
                                        & Test                                & 1,129           & 3,291          & 477            & 559              \\ \hline
\multirow{3}{*}{ACE05-CN}               & Train                               & 6,841           & 29,657         & 7,934          & 2,926            \\ \cline{2-6} 
                                        & Dev                                 & 526             & 2,250          & 596            & 217              \\ \cline{2-6} 
                                        & Test                                & 547             & 2,388          & 672            & 190              \\ \hline
\multirow{3}{*}{ERE-ES}                 & Train                               & 7,067           & 11,839         & 1,698          & 3,272            \\ \cline{2-6} 
                                        & Dev                                 & 556             & 886            & 120            & 210              \\ \cline{2-6} 
                                        & Test                                & 546             & 811            & 108            & 269              \\ \hline
\end{tabular}
}
\caption{Numbers of sentences (i.e., \textbf{sents}), entity mentions (i.e., \textbf{ents}), relations (i.e., \textbf{rels}), and events (i.e., \textbf{events}) in the datasets.}
\label{tab:datasets}
\end{table}




For ERE, following \cite{lin2020joint}, we combine the data from three datasets for English (i.e.,  LDC2015E29, LDC2015E68, and LDC2015E78) that are created under the Deep Exploration and Filtering of Test (DEFT) program (called {\bf ERE-EN}). Similar to ACE05-E+, ERE-EN is also used to evaluate the joint models on four IE tasks.

To demonstrate the portability of our model to other languages, we also apply FourIE to the joint four-IE datasets on Chinese and Spanish. Following \cite{lin2020joint}, we use the ACE 2005 dataset for the evaluation on Chinese (called {\bf ACE05-CN}) and the ERE dataset (LDC2015E107) for Spanish (called {\bf ERE-ES}).


To ensure a fair comparison, we adopt the same data pre-processing and splits (train/dev/test) in prior work \cite{lin2020joint} for all the datasets. As such, ACE05-R, ACE05-E, ACE05-E+, and AC05-CN involve 7 entity types, 6 relation types, 33 event types, and 22 argument roles while ERE-ES and ERE-EN include 7 entity types, 5 relation types, 38 event types, and 20 argument roles. The statistics for the datasets are shown in Table \ref{tab:datasets}.

\noindent \textbf{Hyper-parameters and Evaluation Criteria}. We fine-tune the hyper-parameters for our model using the development data. The suggested values are shown in the appendix. To achieve a fair comparison with \cite{lin2020joint}, we employ the \textit{bert-large-cased} model for the English datasets and \textit{bert-multilingual-cased} model for the Chinese and Spanish datasets. Finally, we follow the same evaluation script and correctness criteria for entity mentions, event triggers, relations, and argument as in prior work \cite{lin2020joint}. The reported results are the average performance of 5 model runs using different random seeds.

\noindent \textbf{Performance Comparison}. We compare the proposed model {\bf FourIE} with two prior models for joint four-task IE: (i) {\bf DyGIE++} \cite{wadden2019entity}: a BERT-based model with span graph propagation, and (ii) {\bf OneIE} \cite{lin2020joint}: {\it the current state-of-the-art (SOTA) model} for joint four-task IE based on BERT and type dependency constraint at the decoding step. Table \ref{tab:main-results} presents the performance (F1 scores) of the models on the test data of the English datasets. Note that in the tables, the prefixes ``Ent'', ``Trg'', ``Rel'', and ``Arg'' represent the extraction tasks for entity mentions, event triggers, relations, and arguments respectively while the suffixes ``-I'' and ``-C'' correspond to the identification performance (only concerning the offset correctness) and identification+classification performance (evaluating both offsets and types).

\begin{table}[ht]
\centering
\small
\resizebox{0.485\textwidth}{!}{
\begin{tabular}{|c|l|c|c|c|c|}
\hline
\textbf{Datasets}         & \multicolumn{1}{c|}{\textbf{Task}} & \textbf{DyGIE++} & \textbf{OneIE} & \textbf{FourIE} & $\Delta\%$    \\ \hline
\multirow{2}{*}{ACE05-R}  & Ent-C                              & 88.6             & 88.8           & \textbf{88.9}   & 0.1  \\ \cline{2-6} 
                          & Rel-C                              & 63.4             & 67.5           & \textbf{68.9}$\dag$   & 1.4  \\ \hline
\multirow{5}{*}{ACE05-E}  & Ent-C                              & 89.7             & 90.2           & \textbf{91.3}$\dag$   & 1.1  \\ \cline{2-6} 
                          & Trg-I                              & -                & 78.2           & \textbf{78.3}   & 0.1  \\ \cline{2-6} 
                          & Trg-C                              & 69.7             & 74.7           & \textbf{75.4}$\dag$   & 0.7  \\ \cline{2-6} 
                          & Arg-I                              & 53.0             & 59.2           & \textbf{60.7}$\dag$   & 1.5  \\ \cline{2-6} 
                          & Arg-C                              & 48.8             & 56.8           & \textbf{58.0}$\dag$   & 1.2  \\ \hline
\multirow{6}{*}{ACE05-E+} & Ent-C                              & -                & 89.6           & \textbf{91.1}$\dag$   & 1.5  \\ \cline{2-6} 
                          & Rel-C                              & -                & 58.6           & \textbf{63.6}$\dag$   & 5.0  \\ \cline{2-6} 
                          & Trg-I                              & -                & 75.6           & \textbf{76.7}$\dag$   & 1.1  \\ \cline{2-6} 
                          & Trg-C                              & -                & 72.8           & \textbf{73.3}$\dag$   & 0.5  \\ \cline{2-6} 
                          & Arg-I                              & -                & 57.3           & \textbf{59.5}$\dag$   & 2.2  \\ \cline{2-6} 
                          & Arg-C                              & -                & 54.8           & \textbf{57.5}$\dag$   & 2.7  \\ \hline
\multirow{6}{*}{ERE-EN}   & Ent-C                              & -                & 87.0           & \textbf{87.4}   & 0.4  \\ \cline{2-6} 
                          & Rel-C                              & -                & 53.2           & \textbf{56.1}$\dag$   & 2.9  \\ \cline{2-6} 
                          & Trg-I                              & -                & 68.4           & \textbf{69.3}$\dag$   & 0.9  \\ \cline{2-6} 
                          & Trg-C                              & -                & 57.0           & \textbf{57.9}$\dag$   & 0.9  \\ \cline{2-6} 
                          & Arg-I                              & -                & 50.1           & \textbf{52.2}$\dag$   & 2.1  \\ \cline{2-6} 
                          & Arg-C                              & -                & 46.5           & \textbf{48.6}$\dag$   & 2.1 \\ \hline
\end{tabular}
}
\caption{F1 scores of the models on the test data of English datasets. $\Delta$ indicates the performance difference between FourIE and OneIE. Rows with $\dag$ designate the significant improvement ($p<0.01$) of FourIE over OneIE.} 
\label{tab:main-results}
\end{table}

As can be seen from the table, FourIE is consistently better than the two baseline models (DyGIE++ and OneIE) across different datasets and tasks. The performance improvement is significant for almost all the cases and clearly demonstrates the effectiveness of the proposed model.



Finally, Table \ref{tab:multilingual} reports the performance of FourIE and OneIE on the Chinese and Spanish datasets (i.e., ACE05-CN and ERE-ES). In addition to the monolingual setting (i.e., trained and evaluated on the same languages), following \cite{lin2020joint}, we also evaluate the models on the multilingual training settings where ACE05-CN and ERE-ES are combined with their corresponding English datasets ACE05-E+ and EAE-EN (respectively) to train the models (for the four IE tasks), and the performance is then evaluated on the test sets of the corresponding languages (i.e., ACE05-CN and ERE-ES). It is clear from the table that FourIE also significantly outperforms OneIE across nearly all the different setting combinations for languages, datasets and tasks. This further illustrates the portability of FourIE to different languages.


\begin{table}[ht]
\centering
\small
\resizebox{0.485\textwidth}{!}{
\begin{tabular}{|c|c|c|c|c|c|}
\hline
\textbf{\begin{tabular}[c]{@{}c@{}}Test Data\end{tabular}} & \textbf{\begin{tabular}[c]{@{}c@{}}Train Data\end{tabular}}                & \textbf{Task} & \textbf{OneIE} & \textbf{FourIE} & $\Delta\%$ \\ \hline
\multirow{8}{*}{ACE05-CN}                                    & \multirow{4}{*}{ACE05-CN}                                                    & Ent-C         & 88.5           & \textbf{88.7}   & 0.2   \\ \cline{3-6} 
                                                             &                                                                              & Rel-C         & 62.4           & \textbf{65.1}$\dag$   & 2.7   \\ \cline{3-6} 
                                                             &                                                                              & Trg-C         & 65.6           & \textbf{66.5}$\dag$   & 0.9   \\ \cline{3-6} 
                                                             &                                                                              & Arg-C         & 52.0           & \textbf{54.9}$\dag$   & 2.9   \\ \cline{2-6} 
                                                             & \multirow{4}{*}{\begin{tabular}[c]{@{}c@{}}ACE05-CN\\ ACE05-E+\end{tabular}} & Ent-C         & \textbf{89.8}           & 89.1   & -0.7  \\ \cline{3-6} 
                                                             &                                                                              & Rel-C         & 62.9           & \textbf{65.9}$\dag$   & 3.0   \\ \cline{3-6} 
                                                             &                                                                              & Trg-C         & 67.7           & \textbf{70.3}$\dag$   & 2.6   \\ \cline{3-6} 
                                                             &                                                                              & Arg-C         & 53.2           & \textbf{56.1}$\dag$   & 2.9   \\ \hline
\multirow{8}{*}{ERE-ES}                                      & \multirow{4}{*}{ERE-ES}                                                      & Ent-C         & 81.3           & \textbf{82.2}$\dag$   & 0.9   \\ \cline{3-6} 
                                                             &                                                                              & Rel-C         & 48.1           & \textbf{57.9}$\dag$   & 9.8   \\ \cline{3-6} 
                                                             &                                                                              & Trg-C         & 56.8           & \textbf{57.1}   & 0.3   \\ \cline{3-6} 
                                                             &                                                                              & Arg-C         & 40.3           & \textbf{42.3}$\dag$   & 2.0   \\ \cline{2-6} 
                                                             & \multirow{4}{*}{\begin{tabular}[c]{@{}c@{}}ERE-ES\\ ERE-EN\end{tabular}}     & Ent-C         & 81.8           & \textbf{82.7}$\dag$   & 0.9   \\ \cline{3-6} 
                                                             &                                                                              & Rel-C         & 52.9           & \textbf{59.1}$\dag$   & 6.2   \\ \cline{3-6} 
                                                             &                                                                              & Trg-C         & 59.1           & \textbf{61.3}$\dag$   & 2.2   \\ \cline{3-6} 
                                                             &                                                                              & Arg-C         & 42.3           & \textbf{45.4}$\dag$   & 3.1   \\ \hline
\end{tabular}
}
\caption{F1 scores on Chinese and Spanish test sets. $\dag$ marks the significant improvement ($p<0.01$) of FourIE over OneIE.}
\label{tab:multilingual}
\end{table}



\noindent \textbf{Effects of } \texttt{GCN}$^{inst}$ and \texttt{GCN}$^{type}$. This section evaluates the contributions of the two important components in our proposed model FourIE, i.e., the instance interaction graph with $\texttt{GCN}^{inst}$ and the type dependency graph with $\texttt{GCN}^{type}$. In particular, we examine the following ablated/varied models for FourIE: (i) ``\textbf{FourIE}-\texttt{GCN}$^{inst}$'': this model excludes the instance interaction graph and the GCN model \texttt{GCN}$^{inst}$ from FourIE so the initial instance representations $\textbf{r}_k$ are directly used to predict the types for the instances (replacing the enriched vectors $\textbf{r}^{inst}_k$), (ii) ``\textbf{FourIE}-\texttt{GCN}$^{type}$'': this model eliminates the type dependency graph and the GCN model \texttt{GCN}$^{type}$ (thus the loss term $L_{dep}$ as well) from FourIE, (iii)  ``\textbf{FourIE}-\texttt{GCN}$^{inst}$-\texttt{GCN}$^{type}$'': this model removes both the instance interaction and type dependency graphs from FourIE, (iv) ``\textbf{FourIE}-\texttt{GCN}$^{type}$+TDDecode'': this model also excludes \texttt{GCN}$^{type}$; however, it additionally applies the global type dependencies features to score the joint predictions for the beam search in the decoding step (the implementation for this beam search is inherited from \cite{lin2020joint} for a fair comparison), and (v) ``\textbf{FourIE}-$\hat{\textbf{A}}^{pred}$'': instead of employing the approximation matrix $\hat{\textbf{A}}^{pred}$ in FourIE, this model directly uses the adjacency matrix $\textbf{A}^{pred}$ in the $L_{dep}$ regularizer ($L_{dep}$ thus does not influence the instance representation-related parameters $\theta^{inst}$). Table \ref{tab:main-ablation} shows the performance of the models on the development dataset of ACE05-E+ for four IE tasks.

\begin{table}[ht]
\centering
\small
\resizebox{0.485\textwidth}{!}{
\begin{tabular}{|l|c|c|c|c|}
\hline
\textbf{Models}      & \textbf{Ent-C} & \textbf{Rel-C} & \textbf{Trg-C} & \textbf{Arg-C} \\ \hline
\textbf{FourIE}               & {\bf 89.6}           & {\bf 64.3}           & {\bf 71.0}           & {\bf 59.0}           \\ \hline \hline
\textbf{FourIE}-\texttt{GCN}$^{inst}$        & 89.1           & 62.3           & 70.3           & 57.5           \\ \hline
\textbf{FourIE}-\texttt{GCN}$^{type}$        & 88.5           & 61.8           & 69.9           & 56.6           \\ \hline
\textbf{FourIE}-\texttt{GCN}$^{inst}$-\texttt{GCN}$^{type}$ & 88.2           & 59.3           & 68.9           & 56.1           \\ \hline
\textbf{FourIE}-\texttt{GCN}$^{type}$+TDDecode       & 88.8           & 59.6           & 70.8           & 56.8           \\ \hline
\textbf{FourIE}-$\hat{\textbf{A}}^{pred}$ & 89.0 & 62.3 & 70.2 & 57.6 \\ \hline 
\end{tabular}
}
\caption{F1 scores of the models on the ACE05-E+ dev data.}
\label{tab:main-ablation}
\end{table}

The most important observation from the table is that both \texttt{GCN}$^{inst}$ and \texttt{GCN}$^{type}$ are necessary for FourIE to achieve the highest performance for the four IE tasks. Importantly, replacing \texttt{GCN}$^{type}$ in FourIE with the global type dependency features for decoding (i.e., ``\textbf{FourIE}-\texttt{GCN}$^{type}$+TDDecode'') as in \cite{lin2020joint} or eliminating the approximation $\hat{\textbf{A}}^{pred}$ for $L_{dep}$ produces inferior performance, especially for relation and argument extraction. This clearly demonstrates the benefits for deeply integrating knowledge from type dependencies to influence representation learning parameters with $L_{dep}$ for joint four-task IE.


\noindent {\bf Contributions of Type Dependency Edges}. Our type dependency graphs $\textbf{G}^{gold}$ and $\textbf{G}^{pred}$ involves three categories of edges, i.e., entity\_relation, entity\_argument, and event\_argument type edges. Table \ref{tab:type-ablation} presents the performance of FourIE (on the development data of ACE05-E+) when each of these edge categories is excluded from our type dependency graph construction.

\begin{table}[ht]
\centering
\small
\resizebox{0.485\textwidth}{!}{
\begin{tabular}{|l|c|c|c|c|}
\hline
\textbf{Models} & \textbf{Ent-C} & \textbf{Rel-C} & \textbf{Trg-C} & \textbf{Arg-C} \\ \hline
FourIE          & {\bf 89.6}           & {\bf 64.3}           & {\bf 71.0}           & {\bf 59.0}           \\ \hline \hline
FourIE - entity\_relation    & 88.7           & 61.9           & 71.0           & 57.5           \\ \hline
FourIE - entity\_argument    & 89.3           & 63.2           & 70.0           & 56.9           \\ \hline
FourIE - event\_argument    & 89.5           & 64.1           & 69.8           & 57.7           \\ \hline
\end{tabular}
}
\caption{F1 scores of the ablated models for type dependency edges on the ACE05-E+ dev data.}
\label{tab:type-ablation}
\end{table}

The table clearly shows the importance of different categories of type dependency edges for FourIE as the elimination of any category would generally hurt the performance of the model. In addition, we see that the contribution level of the type dependency edges intuitively varies according to the tasks of consideration. For instance, entity\_relation type edges are helpful mainly for entity mention, relation and argument extraction. Finally, an error analysis is conducted in the appendix to provide insights about the benefits of the type dependency graphs $\textbf{G}^{gold}$ and $\textbf{G}^{pred}$ for FourIE (i.e., by comparing the outputs of FourIE and ``\textbf{FourIE}-\texttt{GCN}$^{type}$'').

\section{Related Work}

The early joint methods for IE have employed feature engineering to capture the dependencies between IE tasks, including Integer Linear Programming for Global Constraints \cite{Roth:04,Li:11}, Markov Logic Networks \cite{Riedel:09,Venugopal:14}, Structured Perceptron \cite{Li:13,Li:14,Miwa:14,Alex:16}, and Graphical Models \cite{yu2010jointly,Yang:16}.


Recently, the application of deep learning has facilitated the joint modeling for IE via shared parameter mechanisms across tasks. These joint models have focused on different subsets of the IE tasks, including EME and RE \cite{Zheng:17,Katiyar:17,Bekoulis:18,Fu:19,luan2019general,Sun:19,Veyseh:20multiview,Veyseh:20exploiting}, event and temporal RE \cite{Han:19}, and ETD and EAE \cite{Nguyen:16,Zhang:19,Nguyen:19oneforall}. However, none of these work has explored joint inference for four IE tasks EME, ETD, RE, and EAE as we do. The two most related works to ours include \cite{wadden2019entity} that leverages the BERT-based information propagation via dynamic span graphs, and \cite{lin2020joint} that exploits BERT and global type dependency features to constrain the decoding step. Our model is different from these works in that we introduce a novel interaction graph for instance representations for four IE tasks and a global type dependency graph to directly inject the knowledge into the training process.


\section{Conclusion}
We present a novel deep learning framework to jointly solve four IE tasks (EME, ETD, RE, and EAE). Our model attempts to capture the inter-dependencies between instances of the four tasks and their types based on instance interaction and type dependency graphs. GCN models are employed to induce representation vectors to perform type predictions for task instances and regularize the training process. The experiments demonstrate the effectiveness of the proposed model, leading to SOTA performance over multiple datasets on English, Chinese, and Spanish. In the future, we plan to extend the model to include more IE tasks (e.g., coreference resolution).




\section*{Acknowledgments}

This research has been supported by the Army Research Office (ARO) grant W911NF-21-1-0112. This research is also based upon work supported by the Office of the Director of National Intelligence (ODNI), Intelligence Advanced Research Projects Activity (IARPA), via IARPA Contract No. 2019-19051600006 under the Better Extraction from Text Towards Enhanced Retrieval (BETTER) Program. The views and conclusions contained herein are those of the authors and should not be interpreted as necessarily representing the official policies, either expressed or implied, of ARO, ODNI, IARPA, the Department of Defense, or the U.S. Government. The U.S. Government is authorized to reproduce and distribute reprints for governmental purposes notwithstanding any copyright annotation therein. This document does not contain technology or technical data controlled under either the U.S. International Traffic in Arms Regulations or the U.S. Export Administration Regulations.

\bibliography{anthology,custom}
\bibliographystyle{acl_natbib}

\clearpage

\appendix

\section{Hyper-parameters}


We fine-tune the hyper-parameters for our model FourIE using the development data of the ACE05-E+ dataset (our main and largest evaluation dataset). The selection criteria is based on the average F1 scores of the four IE tasks of consideration (EME, ETD, RE, and EAE). The following values are suggested by the fine-tuning: $2e$-$5$ for the learning rate of BertAdam for the optimizer; 10 for the batch size; $N_i = 2$ and $N_t = 3$ for the numbers of layers in the GCN models $\texttt{G}^{inst}$ and $\texttt{G}^{pred}$ respectively; 300 hidden units for all the layers of the feed forward networks GCN models, and type embeddings; $\beta = 100$ for the constant in the approximation $\hat{\textbf{A}}^{pred}$; $\tau = 0.1$ for the temperature parameter; and $\lambda = 0.5$ for the trade-off parameter in the loss function. To achieve a consistency, we apply the same hyper-parameters from this fine-tuning for other datasets.

\section{Analysis}

\noindent \textbf{Analysis}. To better understand the contribution of the knowledge from the type dependency graphs $\textbf{G}^{gold}$ and $\textbf{G}^{pred}$ for FourIE on EAE, we analyze the set of all the argument instances on the ACE05-E+ development set (called $\mathcal{A}$) that FourIE can successfully predict the argument roles while ``FourIE-\texttt{GCN}$^{type}$'' fails to do so. In particular, we find three major categories of the instances in $\mathcal{A}$ that highlight the benefits of the type dependency graphs:

(i) One-edge constraints (accounting for 28.9\% of $\mathcal{A}$): The incorrect argument role predictions of ``FourIE-\texttt{GCN}$^{type}$'' for these instances violate the constraint on the possible argument roles of event types. As FourIE does not have this issue, it suggests that FourIE can learn and enforce those constraints (i.e., from the event\_argument edges of $\textbf{G}^{gold}$) from the training. For instance, in the sentence ``{\it ... the United States upped its military presence, \underline{deploying} more missile-firing warships to the \textbf{Red Sea}}'', both FourIE and ``FourIE-\texttt{GCN}$^{type}$'' can recognize ``{\it deploying}'' as an event trigger of type {\it Transport}. However, regarding the entity mention ``{\it Red Sea}'', FourIE correctly assigns the {\it Destination} role for the {\it Transport} event while ``FourIE-\texttt{GCN}$^{type}$'' incorrectly considers it as the role {\it Place} (an invalid role for the event type {\it Transport}).

(ii) Two-edge constraints (representing 36.5\% of $\mathcal{A}$): The predictions from ``FourIE-\texttt{GCN}$^{type}$'' in this category involve argument roles that are never assigned to an entity mention of some entity type in an event mention/trigger of some event type. FourIE can avoid this issue as it can recognize these constraints from the combinations of two neighboring edges (i.e., an event\_argument and and entity\_argument edge). For example, in the sentence ``\textit{... the tanks and Bradley fighting vehicles ... backed by the Apache attack \textbf{helicopters} ... \underline{punched through} the Republican Guard defenses ...}'', both FourIE and ``FourIE-\texttt{GCN}$^{type}$'' can detect ``\textit{helicopters}'' as an entity mention of type \textit{Vehicle} which is an argument for the ``\textit{Attack}'' event triggered by ``\textit{punched through}''. However, ``FourIE-\texttt{GCN}$^{type}$'' incorrectly predicts the argument role of \textit{``Attacker''} for \textit{``helicopters''} while FourIE can successfully return the {\it Instrument} in this case. In fact, we cannot find any {\it Vehicle} entity that plays the {\it Attacker} role in an {\it Attack} event in the training data, providing an useful information for FourIE to learn and fix the error.


(iii) Four-edge constraints (accounting for 19.2\% of $\mathcal{A}$): The failure of ``FourIE-\texttt{GCN}$^{type}$'' for the instances in this category can be fixed if the model exploits the co-occurrence of event types and argument roles in the same sentences. In particular, for two event mentions with related event types in the same sentences, an entity mention that plays some role in one event tends to also play some related role in the other event. These co-occurrence can be captured via two event\_argument edges and two entity\_argument edges (sharing the same entity type) in the type dependency graphs of FourIE to address the issue. Consider an example sentence: ``{\it Two 13-year-old children were among \textbf{those} \underline{killed} in the Haifa bus \underline{bombing}, Israeli public radio said ...}''. Both FourIE and ``FourIE-\texttt{GCN}$^{type}$'' can identify the {\it Person} entity mention ``\textit{those}'' as the argument of role {\it Victim} for the {\it Die} event triggered by ``{\it killed}''. However, regarding the {\it Attack} event triggered by ``{\it bombing}'', only FourIE can correctly predict ``\textit{those}'' as an argument of role {\it Target}''. This success can be attributed to the ability of FourIE to learn the co-occurrence that an entity mention has a higher chance to play the role {\it Target} in an {\it Attack} event if it also has a role of {\it Victim} for a {\it Die} event mentioned in the same sentence.

Finally, the instances in the remaining 15.4\% of $\mathcal{A}$ tend to involve more complicated constraints/dependencies that cannot be associated with any of the three categories above.

\end{document}